\documentclass[10pt,twocolumn,letterpaper]{article}

\usepackage[pagenumbers]{cvpr} 

\usepackage{graphicx}
\usepackage{amsmath}
\usepackage{amssymb}
\usepackage{booktabs}
\usepackage{subfloat}  

\usepackage[pagebackref,breaklinks,colorlinks]{hyperref}

\usepackage[capitalize]{cleveref}
\crefname{section}{Sec.}{Secs.}
\Crefname{section}{Section}{Sections}
\Crefname{table}{Table}{Tables}
\crefname{table}{Tab.}{Tabs.}

\def\cvprPaperID{8977} 
\def\confName{CVPR}
\def\confYear{2022}

\begin{document}

\title{$\ell$-Leaks:Membership Inference Attacks with Logits}

\author{Shuhao Li
\and
Yajie Wang\\
\and
Yuan Tan\\
\and
Yuanzhang Li
}
\maketitle

\begin{abstract}
  Machine Learning (ML) has made unprecedented progress in the past several decades. However, due to the memorability of the training data, ML is susceptible to various attacks, especially Membership Inference Attacks (MIAs), the objective of which is to infer the model's training data. So far, most of the membership inference attacks against ML classifiers leverage the shadow model with the same structure as the target model. However, empirical results show that these attacks can be easily mitigated if the shadow model is not clear about the network structure of the target model.
   
   In this paper, We present attacks based on black-box access to the target model. We name our attack \textbf{l-Leaks}. The l-Leaks follows the intuition that if an established shadow model is similar enough to the target model, then the adversary can leverage the shadow model's information to predict a target sample's membership.The logits of the trained target model contain valuable sample knowledge. We build the shadow model by learning the logits of the target model and making the shadow model more similar to the target model. Then shadow model will have sufficient confidence in the member samples of the target model. We also discuss the effect of the shadow model's different network structures to attack results. Experiments over different networks and datasets demonstrate that both of our attacks achieve strong performance.
   
\end{abstract}

\section{Introduction}
\label{sec:intro}

Over the past decade, machine learning (ML) has witnessed tremendous progress, enabling a wide range of applications, such as image recognition and machine translation. However, recent work has demonstrated that ML, especially Deep Learning(DL), is vulnerable to privacy leakage since they are shown to overfit their training datasets. Adversaries can exploit this vulnerability with query access to the trained models or the synthetic data outputs\cite{carlini2019secret} \cite{chen2020machine} \cite{ganju2018property} \cite{he2021stealing} \cite{melis2019exploiting} \cite{pan2020privacy} \cite{salem2020updates}, to infer membership of samples to the training datasets. 
\begin{figure}[ht]
	\centering
	\includegraphics[scale=0.4]{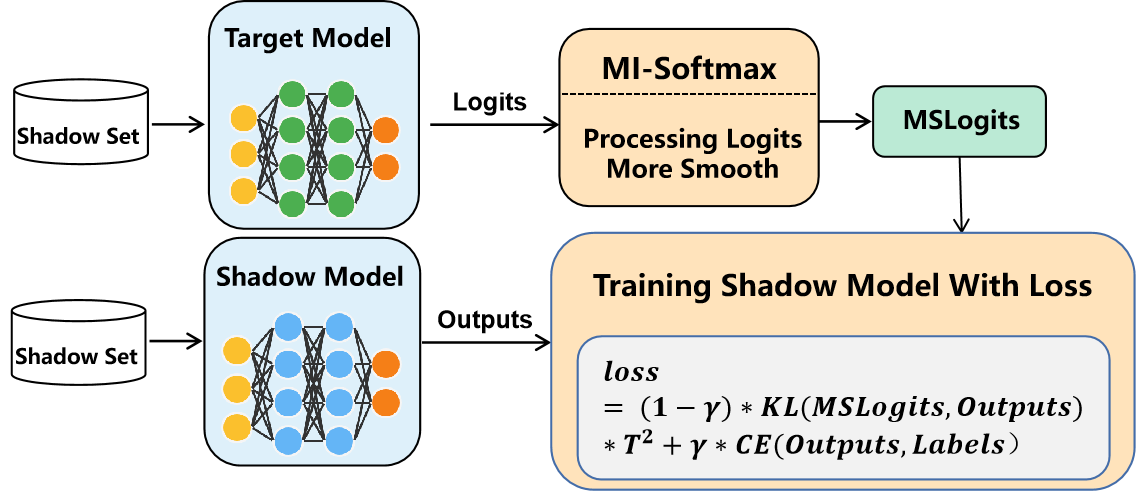}
	\caption{Construction of the Shadow model. Input $D_{shadow}$ into shadow model and target model to obtain logits. As shown in Fig.\ref{fig:logitsintro}, the logits of the target model contains valuable sample knowledge. Therefore, we designed \textbf{MI-Softmax} to process the logits of the target model to retain the valuable sample knowledge called MSLogits. The shadow model is constructed by reducing the gap between the Outputs and MSLogits.The shadow model will be more similar to the target model by learning the behavior of the target model.}
	\label{fig:trainshadow}
\end{figure}

This paper focuses on Membership Inference Attacks(MIA). Its purpose is to determine whether a data sample is present in the original training set. For example, suppose we use a tumor patient's data in a hospital to train an artificial intelligence system that predicts whether or not an undiagnosed patient has a tumor. In that case, MIA will increase the risk of leaking sensitive patient data. Besides, personal credit, daily physical location, and other data are also vulnerable to MIA.

Membership inference attacks rely on the shadow model to construct an attack model. The responsibility of the shadow model is to imitate the behavior of the target model, so how to build the shadow model directly affects the success rate of the final inference attack.

At present, there is no better way to build a shadow model. For example, in the existing MIA attack, Shokri et al \cite{shokri2017membership} emphasized that the shadow model and the target model have the same network structure in the stage of building the shadow model. It is easy to achieve when the model structure is known, which loosens the shadow model's construction restriction. The sampling attack method of Rahimian et al. \cite{rahimian2020sampling} also uses the same structure as the target model for constructing the shadow model. The Label-Leaks method of Zheng Li et al.\cite{li2020membership} mentioned that When the adversary knows the main task of the target model, it only needs to use a neural network similar to the target model to act as a shadow model. This approach seems to increase the degree of restriction on the shadow model, but there is no theoretical basis for constructing the shadow model by guessing.

Assuming that the structure and parameters of the target model can't be known, we can't set the same model structure as the target model or rely on guessing to develop the shadow model. In that case, the membership inference attacks can't fully play the attack effect in the above paper.
\begin{figure}[ht]
	\centering
	\includegraphics[scale=0.5]{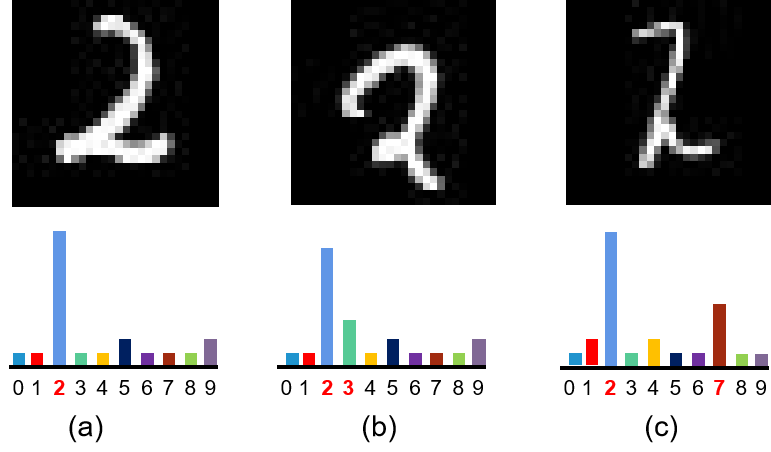}
	\caption{For different samples, logits will contain valuable sample knowledge. For (a), the output is ground truth $2$.For (b), logits will tell us that the high probability of this picture is 2, and the small probability is 3, hardly like other numbers. For (c), the high probability is 2, and the small probability is 7. The information carried in logits will help the shadow model learn the behavior of the target model.}
	\label{fig:logitsintro}
\end{figure}

	In this paper, we propose a new inference attack method, termed \textbf{l-Leaks}. We solve the problem of shadow-model construction. Our intuition is that the logits of the target model provide more supervision signals than one-hot tags.
	
	In machine learning, we often assume a potential functional relationship between the input and the output. This function is unknown: learning a new model is to approximate an unknown function from limited data. Suppose we want the shadow model to be approximately equal to the target model, and the functional of the target model is known. In that case, we can use a lot of pseudo-data within the non-training set to train the shadow model, matching the logits distribution of the shadow model to the real labels. Now,  to make the shadow model more similar to the target model without knowing the structure of the target model network, we just need to make the shadow model and the target model match the logits distribution under the same input. Intuitively, the advantage is that the logits distribution of the trained target model contains valuable sample knowledge. As shown in Fig.\ref{fig:logitsintro}, the actual label can only tell us that Fig.\ref{fig:logitsintro}b is number 2, not 3, and not 7; but the trained logits may tell us that it is most likely to be the number 2, and it is unlikely to be 3, but it can never be 7. The behavior of logits is our intuition to construct a shadow model.
	
	Our ultimate goal is to make the distribution of the softmax output of the shadow model and the target model sufficiently close. However, doing so directly is problematic: in the general softmax function, the natural exponent $e$ will widen the logit gap and normalize. The final distribution approximates $argmax$, and its output is vectors that are close to one-hot. One of the values is very large, and the others are tiny. In this case, the knowledge of "maybe 3, but never 7" mentioned earlier is very limited. 
	
	Compared with the hard output like one-hot, we want the output more smooth, so we designed \textbf{MI-Softmax}(more details in Sec.\ref{sec:methodology}) to make the distribution of the entire logits smoother. As shown in Fig.\ref{fig:trainshadow}, in the shadow model training process, narrow the gap between the output and MSLogits distributions through KL divergence. During the training process, the cross-entropy of the shadow set label and its softmax distribution is added to the loss function to add more supervision signals, which will significantly improve the previous shadow model. It will improve the phenomenon that the shadow model is "biased" by the target model, and the shadow model will have a better-imitating effect on the target model.
	
	The change of l-Leaks lies in constructing the shadow model on a theoretical basis instead of guessing or adopting the same structure as the target model. l-Leaks has a strong performance in our experiments.

We summarize our contributions as below: first, we proposed \textbf{l-Leaks }method, which solves the unfounded problem of the previous shadow-model construction; second, in the process of constructing the shadow model, the concept of \textbf{MI-Softmax} was proposed to solve the problem of model output; finally, Our method was tested on different networks and 5 different datasets. We obtained a high attack success rate. Our approach achieves strong performance in deep neural network scenarios.

\section{Related Work}
\label{sec:relatework}
\subsection{MIAs on Classification Models}

Membership inference attack refers to an attacker trying to determine whether a piece of personal information exists in the training dataset of the target model. When the training data contains sensitive information such as medical data, the data owner does not want to reveal whether the personal data exists in a specific training set. However, the membership inference attack revealed this type of privacy.

Shokri et al. \cite{shokri2017membership} conducted the pioneering work to propose the first MIA on classification models. They invented a shadow training technique to train a binary classifier-based attack model in a black-box setting. Salem et al. \cite{salem2019ml} relax two main assumptions of the shadow training technique in \cite{shokri2017membership}, i.e., multiple shadow models and knowledge of the training data distribution. They argue that the two assumptions are relatively strong, which heavily limit the applicable scenarios of MIAs against ML models. They show that even with one single shadow model, the attacker can achieve comparable attack performance compared to using multiple shadow models. They also propose a data transferring attack where a dataset used to train the shadow model is not required to have the same distribution as the target model’s private training dataset. Yeom et al. \cite{yeom2018privacy} also propose two metric-based MIAs, i.e., the prediction correctness based MIA and the prediction loss based MIA. Compared to binary
classifier-based MIAs, metric-based MIAs are much simpler and cause less computation cost. Long et al. \cite{long2018understanding} \cite{long2020pragmatic} investigate MIAs on ML models which are not overfitted to their training data. They propose a generalized MIA that can identify the membership of, particularly vulnerable records.

\subsection{MIAs based metric}

Li and Zhang \cite{li2020membership} propose two label-only MIAs, i.e., a transfer-based MIA and a perturbed-based MIA. The transfer-based MIA aims to construct a shadow model to mimic the target model. The intuition is that if the shadow model is similar enough to the target model, then the shadow model’s confidence scores on an input record will indicate its membership. The perturbation-based attack adds crafted noise to the target record to turn it into an adversarial example. The intuition is that it is harder to perturb a member instance to a different class than a non-member instance. Thus, the magnitude of the perturbation can be used to distinguish members from non-members. Choquette et al. \cite{choquette2021label} also propose two label-only MIAs, i.e., data augmentation-based MIA and decision boundary distance-based MIA.

\subsection{MIAs on Generative Models}

Although most of the current MIA is carried out for classification models, some MIA efforts are also for generative models.

Hayes et al. \cite{hayes2019logan} introduce the first MIA on generative models in both black-box and white-box settings.Hilprecht et al. \cite{hilprecht2019monte} propose two MIAs on generative models. One is a Monte Carlo integration attack designed for GANs in the black-box setting, and the other is a reconstruction attack designed for VAEs in the white-box setting. Liu et al. \cite{liu2019performing} proposed attack begins with attacking a single target record and then extends to a set of records.
\newcommand{\tabincell}[2]{\begin{tabular}{@{}#1@{}}#2\end{tabular}} 
\renewcommand\arraystretch{1.0}
\begin{table*}
	\centering
	
	\begin{tabular}{cccccc}
		\toprule[1.5pt]
		\tabincell{l}{Attack category}	      & Attacks & \tabincell{c}{Training data \\ details} & \tabincell{c}{Target model \\ structure}
		& Predicted label & \tabincell{c}{Detail prediction model \\(e.g. probabilities or logits)}  \\
		
		\midrule
		Prior Works		&\cite{hui2021practical},\cite{song2021systematic},\cite{shokri2021privacy},\cite{leino2020stolen},\cite{rezaei2021difficulty}    & $\surd$ or --     & $\surd$ or --     & $\surd$     & $\surd$     \\

		l-Leaks   & --    & --     & --     & --     & $\surd$     \\
		\bottomrule[1.2pt]
	\end{tabular}
	\caption{Summarize the degree of understanding of different types of member inference attacks on the target model, "$\surd$"means that the attack requires this information,"--" means that this information is not required.}
	\label{zongshu}
\end{table*}
\begin{figure}[ht]
	\centering
	\includegraphics[scale=0.44]{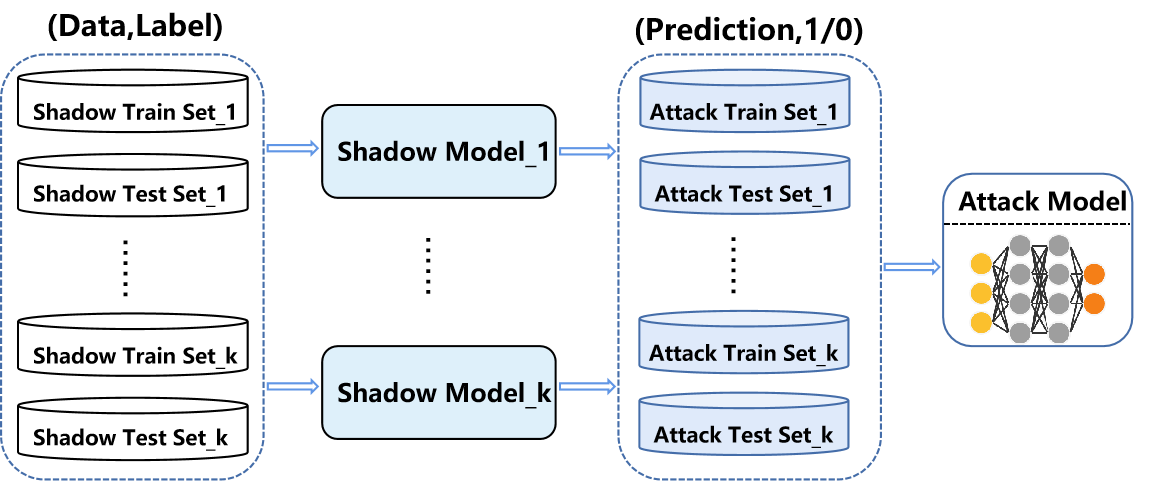}
	\caption{We are training the attack model on the inputs and outputs of the shadow models. For all records in$D_{shadow}^{train}$, we query the shadow model and obtain the output.Then labeled $\textbf{1}$ and added to the $A_{train}$.We also query the shadow model with $D_{test}$ disjoint from $D_{shadow}^{train}$.The outputs on this set are labeled $\textbf{0}$ and also added to $A_{train}$. Having constructed a dataset that reflects the black-box behavior of the shadow models on their training and test datasets, we train a collection of $c_{target}$ attack models, one per output class of the target model.}
	\label{fig:MIA}
\end{figure}
\section{Preliminaries}
\label{sec:relatework}
In this section, we first define the membership inference attack in the machine learning setting, second, describe the metric used, and third, introduce the threat model used for l-Leaks.
\subsection{MIA Against Machine Learning Models}
In this part,we define the membership inference attack in the machine learning setting. We mainly focus on classification tasks in machine learning, one of the most common ML tasks. we use CNN to complete the image classification task, and the selection of this network followed the setting of the previous work \cite{nasr2018machine} \cite{nasr2019comprehensive} \cite{salem2019ml} \cite{shokri2017membership}.A supervised ML classfication model aims to learn a general rule that maps inputs to outputs from a labeled dataset\cite{russell2002artificial}.We supposed $D_{train} = {(x^{(n)},y^{(n)})}_{n=1}^{N}$ be a training dataset,$N$ is the number of sample,$x$ is a feature vector,$y$ is  labels. 

ML model is a function $f(\textbf{x};\Theta)$ ,input $\textbf{x}$ and get output $y=f(\textbf{x};\Theta)$.$\Theta$ is the parameter learned through the training set.When $y$ is a discrete value,  $f(\textbf{x};\Theta)$ is called classification model.

In black-box attack,the attacker intelligently obtains the probability output $p$ by input the data $x$ into the $f(\textbf{x};\Theta)$,then put the obtained probability $p$ into the binary attack model. The attack model will give $0/1$, $0$ means $x$ is not in the training dataset $D_{train}$ and $1$ means $x$ in $D_{train}$.Attack model can be defined as the following function:
$$\kappa = x,f\rightarrow{0,1}$$

$\kappa$ means attack model, $x$ is the input data, $f$ is the target model.
\subsection{Metrics Description}
Our current work is oriented to the classification model in ML. We use \textbf{Accuracy} as a measurement of the target model and shadow model, which is a metric used in existing work to reflect the classification performance of the target model; in addition to the general metrics for evaluating the target model, existing work has also proposed or used many metrics for measuring attack performance. What we use is \textbf{AP}\cite{chen2020practical} \cite{hilprecht2019monte} \cite{liu2021encodermi} \cite{miao2021audio} \cite{zhang2021understanding}, \textbf{AR}\cite{nasr2021adversary} \cite{miao2021audio} \cite{zhang2021understanding}, and \textbf{F1-Score} \cite{he2020segmentations} \cite{hui2021practical} \cite{wu2020characterizing} \cite{zhang2020gan} . A brief introduction to each standard.

\textbf{AP} is what fraction of records classified as members are indeed members of the training dataset.

\textbf{AR} is what fraction of the training dataset’s members are correctly classified as
members.

\textbf{Attack F1-score} is the harmonic mean of Attack Precision and Attack Recall.
\subsection{Threat Model Description}
As shown in Tab.\ref{zongshu}, the adversary can only access the target model through the black box. The information obtained by the attacker is limited to the training data distribution and the black box query on the target model. When the attacker queries the target classifier, only the logits of the input records are obtained. Therefore, the adversary trained a shadow model $S$ to imitate the behavior of the target model $f$, the data for training the shadow model is $D_{shadow}^{train}$, relying on the $S$ to build a dataset for the attack model($\kappa$) and then complete the attack.

The attack process is: first, input $D_{shadow}^{train}$ into $f$ to obtain logits, after that, the obtained logits need to be processed, and the output distribution can be made smoother through MI-Softmax; second, construct shadow model $S$, combine the processed logits and the output of $S$ into a new loss. Step 2 is critical in l-Leaks. It is used for $S$ to learn the behavior of $f$; finally, as shown in Fig.\ref{fig:MIA}, make $S$ construct the training data of the attack model $\kappa$, the training set of $\kappa$ is complete.
\section{Methodology}
\label{sec:methodology}
To solve the problem of shadow model construction in MIA, we propose a model learning construction method based on the target model logits, which achieve the behavior of the shadow model imitating the target model. We provide a theoretical basis for the construction of shadow training in the black box state. Our proposed method aims at the classification task in machine learning. The detailed description of the specific process is as follows.
\subsection{Process logits to Construct Loss Function}
To solve the problem of shadow model construction, we have deeply studied the learning process of the machine learning model. For neural networks, when the input information starts from the input layer (the input layer does not participate in the calculation), the neurons in each layer calculate the output of each neuron in the layer and pass it to the next layer until the output layer calculates the output of the network, feedforward is only used to calculate the output of the network, it does not adjust the parameters of the network. 

Backward propagation is used to adjust the network weights and thresholds during training. The feedback mechanism is used to find the partial derivative, and then the gradient descent is used to find the minimum value of the cost function. This process requires supervised learning. So we use the logits distribution of the model. After the transformation, it matches the label. To construct the shadow model, we need to match the logits distribution of the shadow model and the target model under a given input.

The advantage of constructing a shadow model by learning is that the target model after model training contains valuable sample knowledge in its logits. As shown in Fig.\ref{fig:logitsintro}, ground truth will tell us that Fig.\ref{fig:logitsintro}b is the number $2$; however, logits will tell us that it may be the number $2$, which is unlikely to be the number $3$ and cannot be the number $7$, so we use softmax as shown in Eq.(\ref{con:softmax}) to normalize the logits of the target and shadow models, then match their distribution.

\begin{equation}
	P(z_i)=\dfrac{e^{z_i}}{\sum_{k=1}^{n}{e^{z_i}}}
	\label{con:softmax}
\end{equation}

In subsequent experiments, we found that in the softmax function of Eq.(\ref{con:softmax}), the natural exponent e first widens the gap between logits and then normalizes. The final distribution approximates $argmax$, and its output is a vector close to one-hot. One of the values is very large, and the others are tiny. In this case, the knowledge we need will be weakened, so we hope that after softmax normalizes logits, the information in the data can be retained.

We found that Boltzmann proposed the Softmax function in his foundational statistical mechanic's paper on Boltzmann distribution. The Boltzmann distribution represents a process of simulated energy convergence to the distribution, and its expression is as Eq.(\ref{con:engry}):

\begin{equation}
	F(state) = exp(-\frac{E}{kT})
	\label{con:engry}
\end{equation}

$E$ is the state energy, $k$ is the Boltzmann constant, and $T$ is the thermodynamic temperature.

In statistics, the Boltzmann distribution gives the probability distribution of a system in a specific state of energy and system temperature:
\begin{equation}
	p_i = \frac{exp(-\frac{z_i}{kT})}{\sum_{j=1}^Nexp(-\frac{z_j}{kT})}
	\label{con:Bosoftmax}
\end{equation}

Where $p_i$ is the probability of state $i$, $z_i$ is the state energy, $k$ is Boltzmann's constant, $T$ is the temperature of the system, and $M$ is the number of states of the system.

We discussed the value of variable $T$, $T$ can take three states:
\begin{enumerate}
	\item [1.] When $T = 1$,the formula represents the softmax function.
	\item [2.] When $T=0$,the probability of the best state is $1$, and the probability of the other states is $0$, that is, the probability of the true category is $1$.
	\item [3.] When $T=+\infty$, it is similar to randomly selected states. That is, the probability of each state appearing is equal. This is the so-called maximum entropy state, which is the most chaotic state.
\end{enumerate}

Finally, get a more generalized softmax function, we call it \textbf{MI-Softmax}, as shown in Eq.(\ref{con:MIsoftmax}):

\begin{equation}
	p_i = \frac{exp(\frac{z_i}{T})}{\sum_{j=1}^Nexp(\frac{z_j}{T})}
	\label{con:MIsoftmax}
\end{equation}

After adding T, we can keep more information after normalization. The following describes the learning process.

\subsection{Model Learning to Construct Shadow Model}
We use KL divergence to reduce the gap between the two distributions, as in the following Eq.(\ref{con:kl}).

\begin{equation}
D_{KL}(p_t,q_s) = \sum_{i =1}^N p_t(x_i)log(\frac{p_t(x_i)}{q_s(x_i)})	
	\label{con:kl}
\end{equation}

We only need to minimize the above function. We use the above function to derive an element $z_i$ in logits:

\begin{equation}
	\begin{aligned}
		\dfrac{\partial D}{\partial z_i} &= \dfrac{1}{T}(q_i - p_i)  \\
		&=  \dfrac{1}{T}(\dfrac{\dfrac{exp(z_i)}{T}}{\sum_j exp(\dfrac{z_j}{T})} - \dfrac{\dfrac{exp(v_i)}{T}}{\sum_j exp(\dfrac{v_j}{T})})
	\end{aligned}
\label{con:grad}
\end{equation}

Assume that all logits are zero-averaged for each sample, that is $\sum_j z_j = \sum_j v_j =0 $, then there is:
\begin{equation}
	\begin{aligned}
		\dfrac{\partial D}{\partial z_i} &=  \dfrac{1}{T}(\dfrac{1 + \dfrac{exp(z_i)}{T}}{N} - \dfrac{1+\dfrac{exp(v_i)}{T}}{N})\\
		&=\dfrac{1}{NT^2}(z_i - v_i)
	\end{aligned}
	\label{con:gradresult}
\end{equation}

Observing Eq.(\ref{con:gradresult}) finds that if $T$ is huge, logits are zero-averaged for all samples, so multiplied by a $T^2$, it makes the gradient of the two terms of the loss function roughly an order of magnitude. 

If the shadow set is labeled data, we can add the label's cross-entropy and the shadow model's softmax distribution to the loss function, as shown in Eq.(\ref{con:loss}). Experiments show that the above operation will significantly improve the performance of the shadow model because the training of the model adds more supervise signals.
\begin{equation}
	\begin{aligned}
		L &= \alpha T^2 L_{KL}(p_t,q_s) + \beta L_{CE}(q_s,label) \\
		&=  \alpha T^2 \sum_{i =1}^N p_t(x_i)log(\frac{p_t(x_i)}{q_s(x_i)}) + \beta \sum_{i =1}^N c_jlog(q_{j}^1)
	\end{aligned}
	\label{con:loss}
\end{equation}

At this point, the construction of the shadow model is completed, and the subsequent attack process is carried out according to Fig.\ref{fig:MIA}.
\section{Experiment}
\label{sec:experiment}
In this section, first, we describe the experimental settings; second, define the dataset used; finally, analyze and evaluate the result of attacks.

\subsection{Evaluation Settings}
The structure of the target model in the experiment uses LeNet-5. The shadow model defaults to a two-layer Cov2d and three-layer fully connected neural network. Finally, we will change the different structures of the shadow model for comparison experiments.

We use Python 3.7 and the PyTorch framework to complete the construction of the attack model. The experiment used is a PC, the GPU is RTX3090, and the video memory is 24G.
\subsection{Dataset Description}
We used two types of data to evaluate our method, binary data and image data. Binary data includes Purchases-100 and Texas-100. Image data includes MNIST, CIFAR-10, and CIFAR-100.

\textbf{Purchase100}. For the Purchase100 dataset contains the shopping records of thousands of online customers. We use a processed and simplified version of this dataset. The dataset contains 600 different products, and each user has a binary form indicating whether she has purchased each product (a total of 19,324 data records). We use 10, 000 randomly selected records from the purchase dataset to train the target model as $D_{target}^{train}$. The rest of the dataset contributes to the test set and the training sets of the shadow models as $D_{shadow}^{train}$.

\textbf{Texas100}. For the Texas-100 dataset is based on the Hospital Discharge Data public use files with information about inpatient stays in several health facilities, released by the Texas Department of State Health Services from 2006 to 2009. The model task is to predict the patient's main procedure based on attributes other than secondary procedures. We focus on the 100 most frequent procedures. The resulting dataset has 67,330 records and 6, 170 binary features. We use 20,000 randomly selected records to train the target model as $D_{target}^{train}$, rest of the data as the training data of shadow model as $D_{shadow}^{train}$.

\textbf{MNIST}. For the MNIST dataset, there are 60,000 training images and a total of 10,000 test images for researchers. We randomly selected 15,000 images in the training set to train the target model as $D_{target}^{train}$, and the remaining images were used to train the shadow model as $D_{shadow}^{train}$.

\textbf{CIFAR}. For the CIFAR-10 and CIFAR-100 datasets, \cite{krizhevsky2009learning}, they both have 60,000 color images.CIFAR-10 is composed of 10 classes. Each class contains 5000 training pictures and 1000 test pictures; CIFAR-100 consists of 100 classes, each class includes 500 training pictures and 100 test pictures. They follow the same usage as MNIST.

\textbf{Attack Set}. The process of generating the attack training set $A_{train}$ is shown in Fig.\ref{fig:MIA},we put $D_{shadow}^{train}$ and $D_{test}$into the shadow model to obtain the output probability $p_{shadow}^{train}$ and $p_{test}$, labeled $(p_{shadow}^{train},1)$ and  $(p_{test},0)$. $0$ means non-shadow model training set. $1$ means it is the shadow model training set. So far, the construction of the attack training set is complete.
\subsection{Attack Result}

\begin{figure*}[htbp]
	\centering
	\subfloat[Attack Precision(AP)]{
    \label{fig:subfig:a}
    \includegraphics[width=0.321\textwidth]{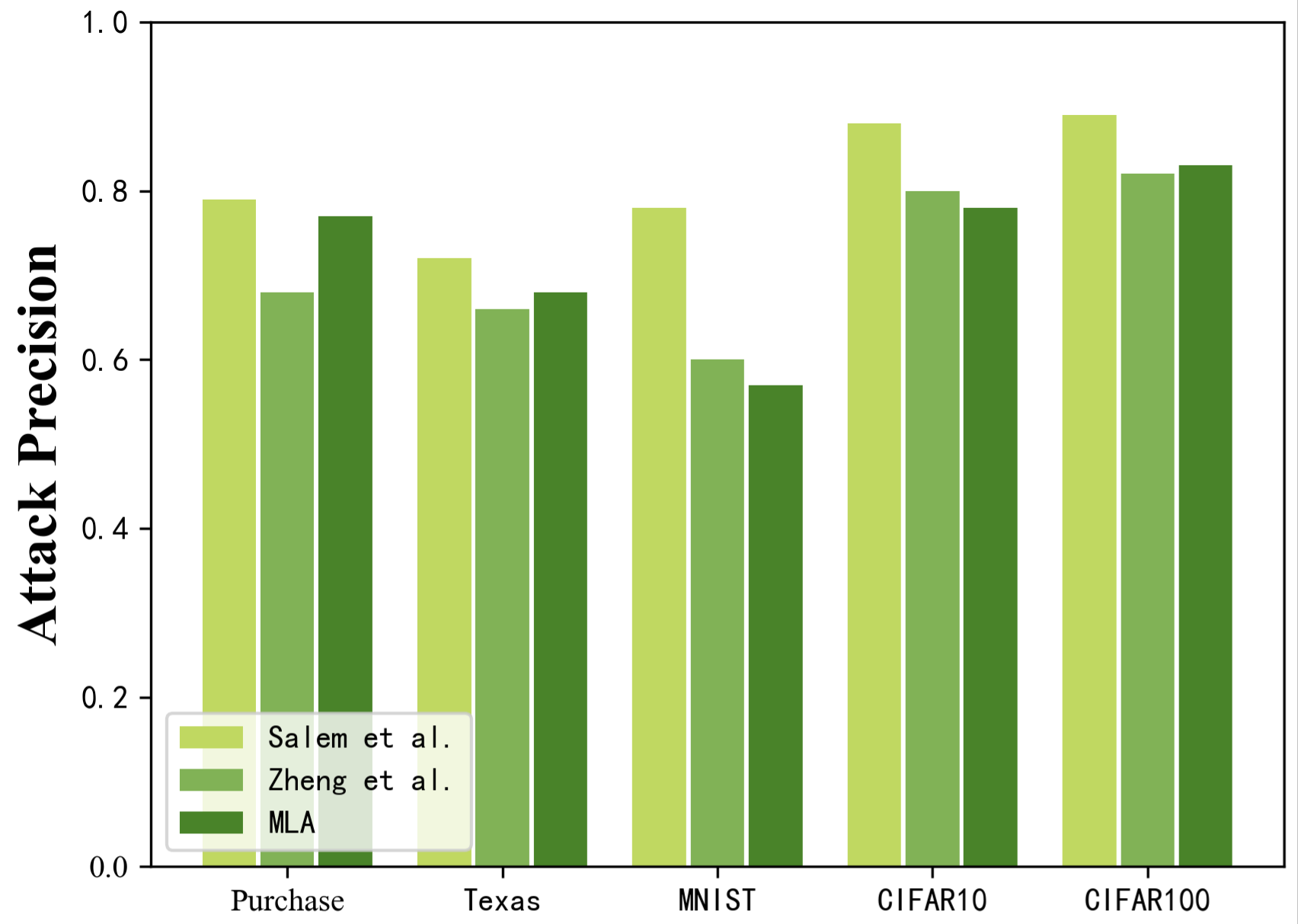}}
	\subfloat[Attack Recall(AR)]{
    \label{fig:subfig:b}
    \includegraphics[width=0.321\textwidth]{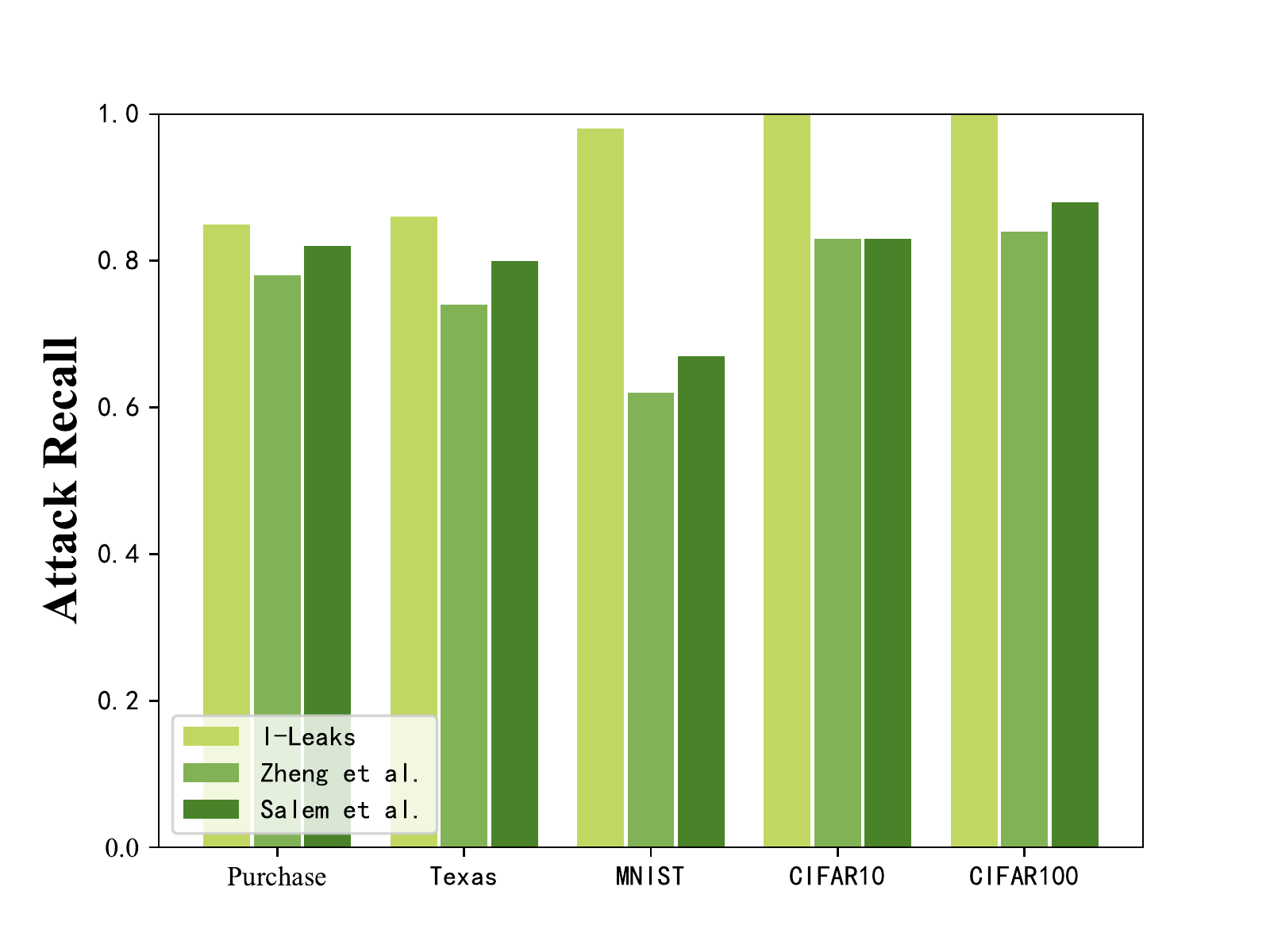}}
    \subfloat[Attack F1-Score]{
    \label{fig:subfig:c}
    \includegraphics[width=0.321\textwidth]{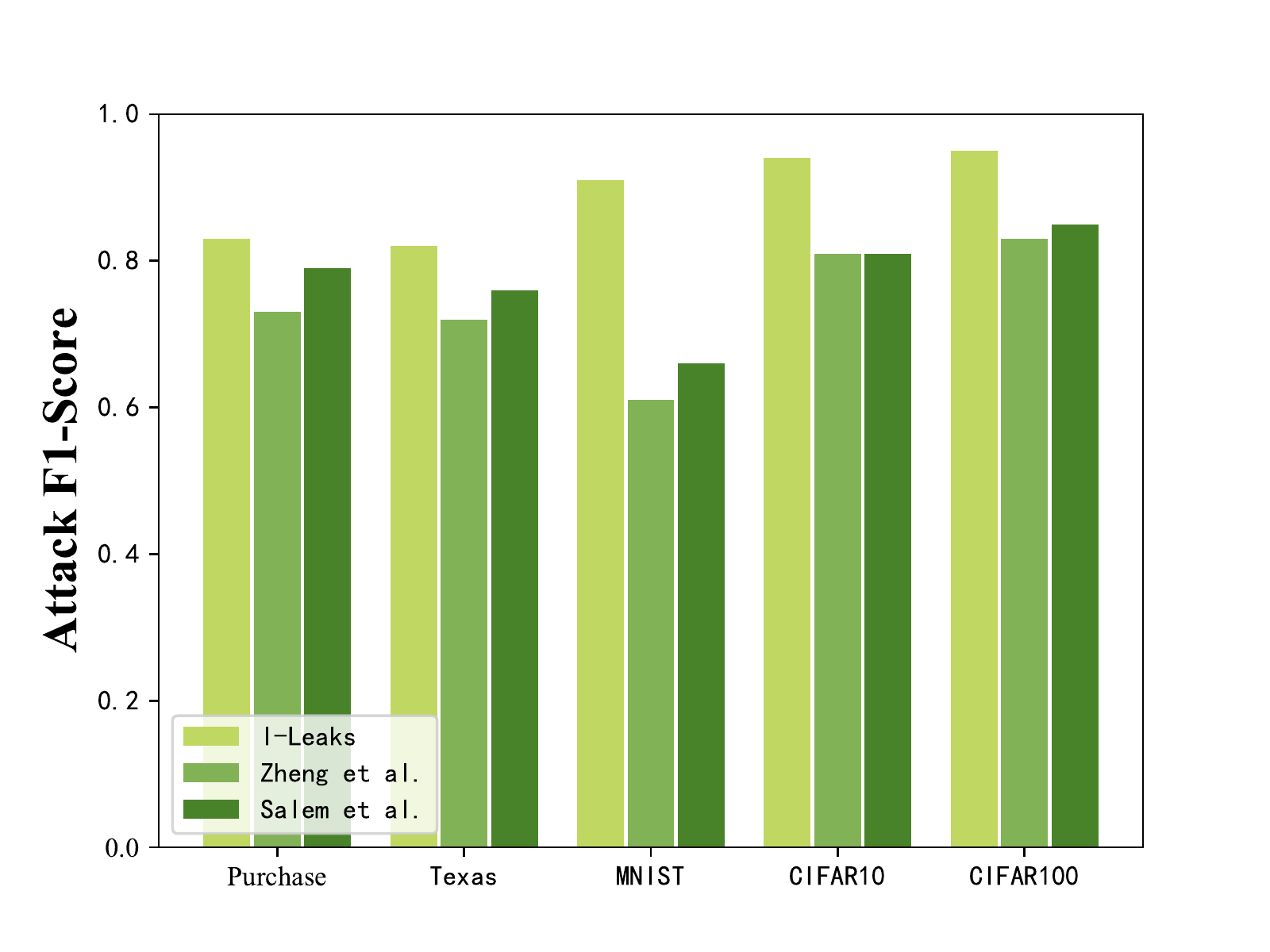}}
	\caption{ Comparison of our l-Leaks performance with the attack by Salem et al and Zheng et al.The metrics we use are \textbf{Attack precision(AP)}, \textbf{Attack Recall(AR)}, and \textbf{Attack f1-score}.The experimental results show that our method has an excellent improvement on the 5 datasets.}
	\label{fig:attackresult}
\end{figure*}

Fig.\ref{fig:attackresult} describes the performance of l-Leaks. We observed that the performance of our attack on the 5 datasets is relatively better than the attacks of Salem et al.\cite{salem2019ml} and Zheng et al.\cite{li2020membership} because our attack uses the logits of the target model as input, the shadow model learns the behavior of the target model, and they are more similar. The shadow model has more confidence in the prediction of the member samples of the target model.

In the work of Salem et al.\cite{salem2019ml}, the shadow model is trained using ground truth labels from the shadow dataset instead of the output of the target model, which shows that the training process is independent of the target model. It also indicates that the closer the overfitting degree of the shadow model and the target model is, the more influential the attack model can be obtained. However, in real applications, the attacker cannot determine the precise overfitting level of the target model, resulting in colossal attack performance wasted. l-Leaks use the logits of the target model to modify the structure of the shadow model and narrow the gap between the posterior distributions of the two models, making the shadow model more similar to the target model. As shown in Fig.\ref{fig:attackresult},we get an excellent attack effect.

\subsection{Evaluation on Shadow Model Performance}
\begin{figure}[h]
	\centering
	\subfloat[Accuracy]{
    \label{fig:acuracy:a}
    \includegraphics[scale=0.28]{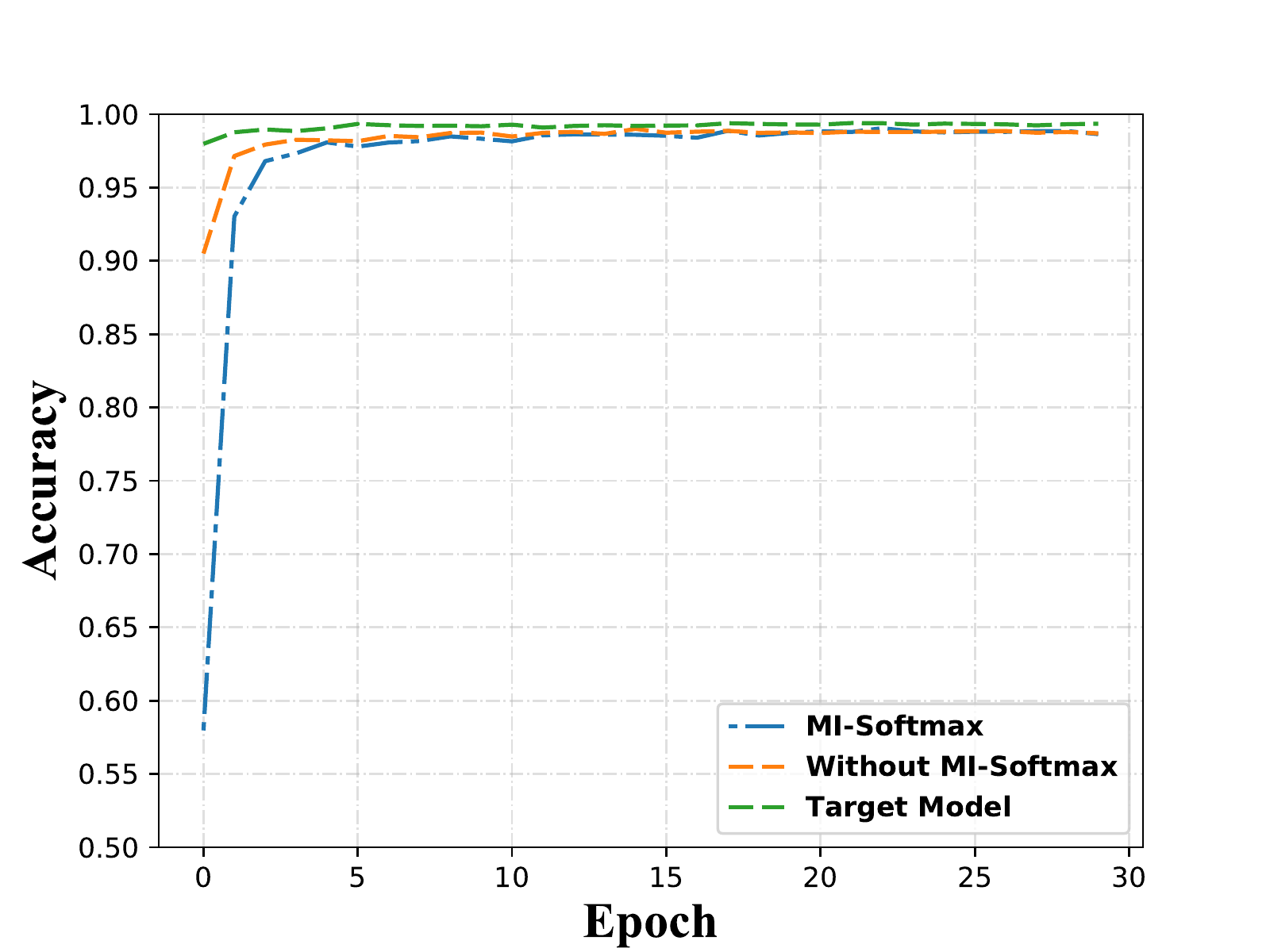}}
    \subfloat[Overfitting Level]{
    \label{fig:accuracy:b}
    \includegraphics[scale=0.275]{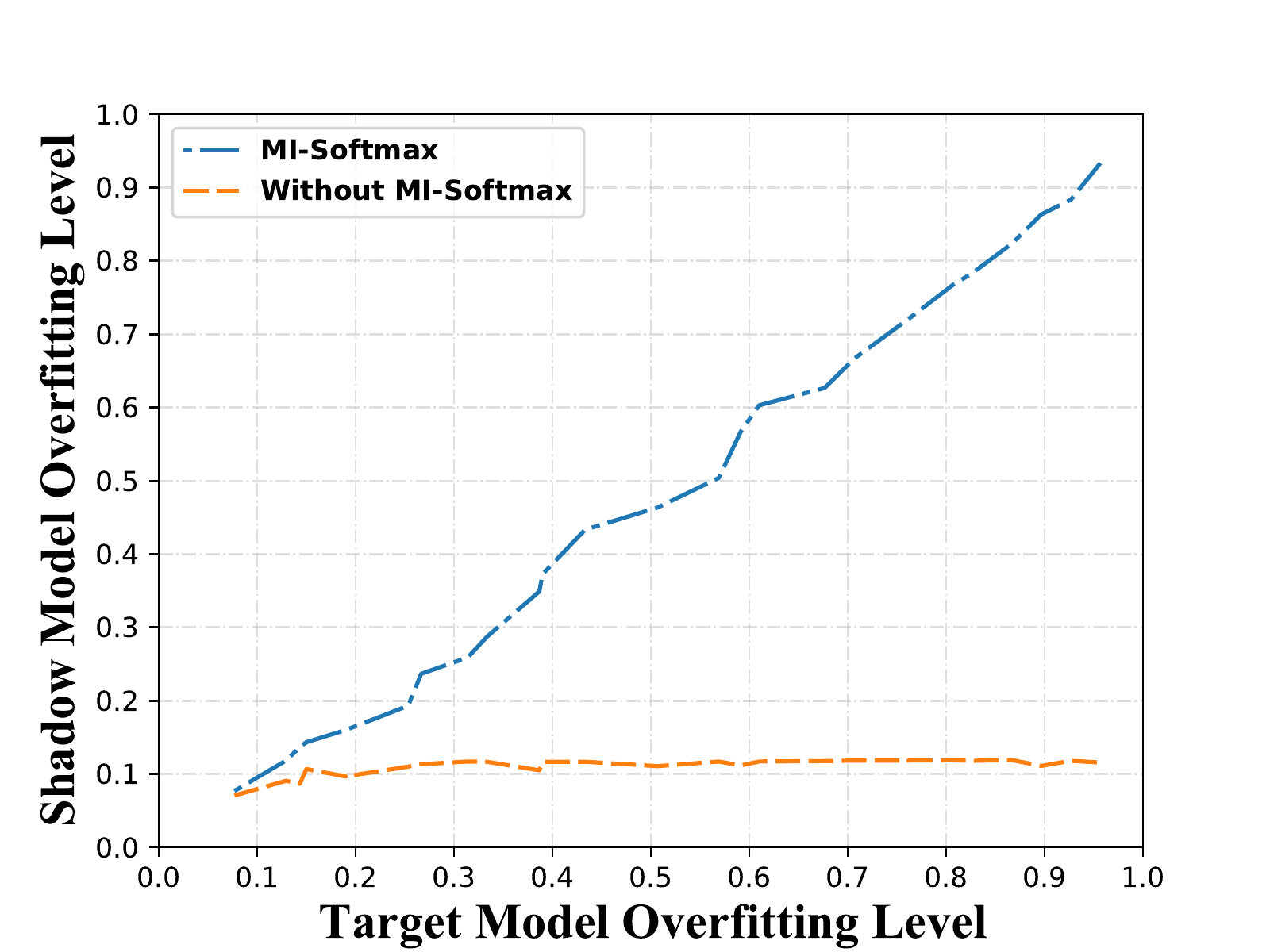}}
	\caption{Shadow model performance comparison. Fig.\ref{fig:acuracy:a} shows the performance of the modify shadow model, original shadow model, and the target model on the test set. Fig.\ref{fig:accuracy:b} shows the relation between the overfitting level of the target model (x-axis) and the performance of the shadow model (y-axis) measured by the difference in prediction accuracy between the $D_{target}^{train}$ and $D_{
test}$.}
	\label{fig:accuracy}
\end{figure}

A major difference between our attack and the previous work is the construction of shadow models. As shown in Fig.\ref{fig:acuracy:a}, we compared the accuracy of our modified shadow model with the accuracy of the original shadow model. Although in the beginning, there was a big gap between our model and the previous model. After 30 epochs, the accuracy of the processed shadow model and the original shadow is \textbf{98.5\%}, \textbf{98.7\%}, the target model is \textbf{99.1\%}, and there is almost no difference between them, indicating that the shadow model maintains the performance of the model after modification.

In Fig.\ref{fig:accuracy:b}.we quantified the \textbf{overfitting level} of the model as the difference in its prediction accuracy on the $D_{target}^{train}$ and $D_{test}$, similar to previous works\cite{salem2019ml}\cite{yeom2018privacy}. We find that the modified shadow model increases with the increase of the overfitting level of the target model, showing a nearly linear relationship. In contrast, the original shadow model has a larger difference, showing the same confidence level for both the $D_{target}^{train}$ and $D_{test}$. This experiment demonstrates that the modified shadow model mimics the behavior of the target model and gets the same performance as the target model on the $D_{target}^{train}$.

\subsection{Evaluation on Shadow Model Similarity}
To prove that the shadow model has learned the behavior of the target model,  we use the previous work\cite{li2020membership} method to verify the performance of the shadow model to imitate the target model. If the shadow model $S$ and the target model $f$ are sufficiently similar, the member state of the data sample in the target model can be reflected in the shadow model. To quantify the prediction confidence of the target sample on the shadow model $S$, we pay attention to the loss of the target sample relative to the shadow model $S$. Note that the target sample is never used to train the shadow model.

\begin{figure*}[h]
\centering
\subfloat[MNIST:Modify Shadow Model]{
\label{fig:subfig:a}
\includegraphics[scale=0.243]{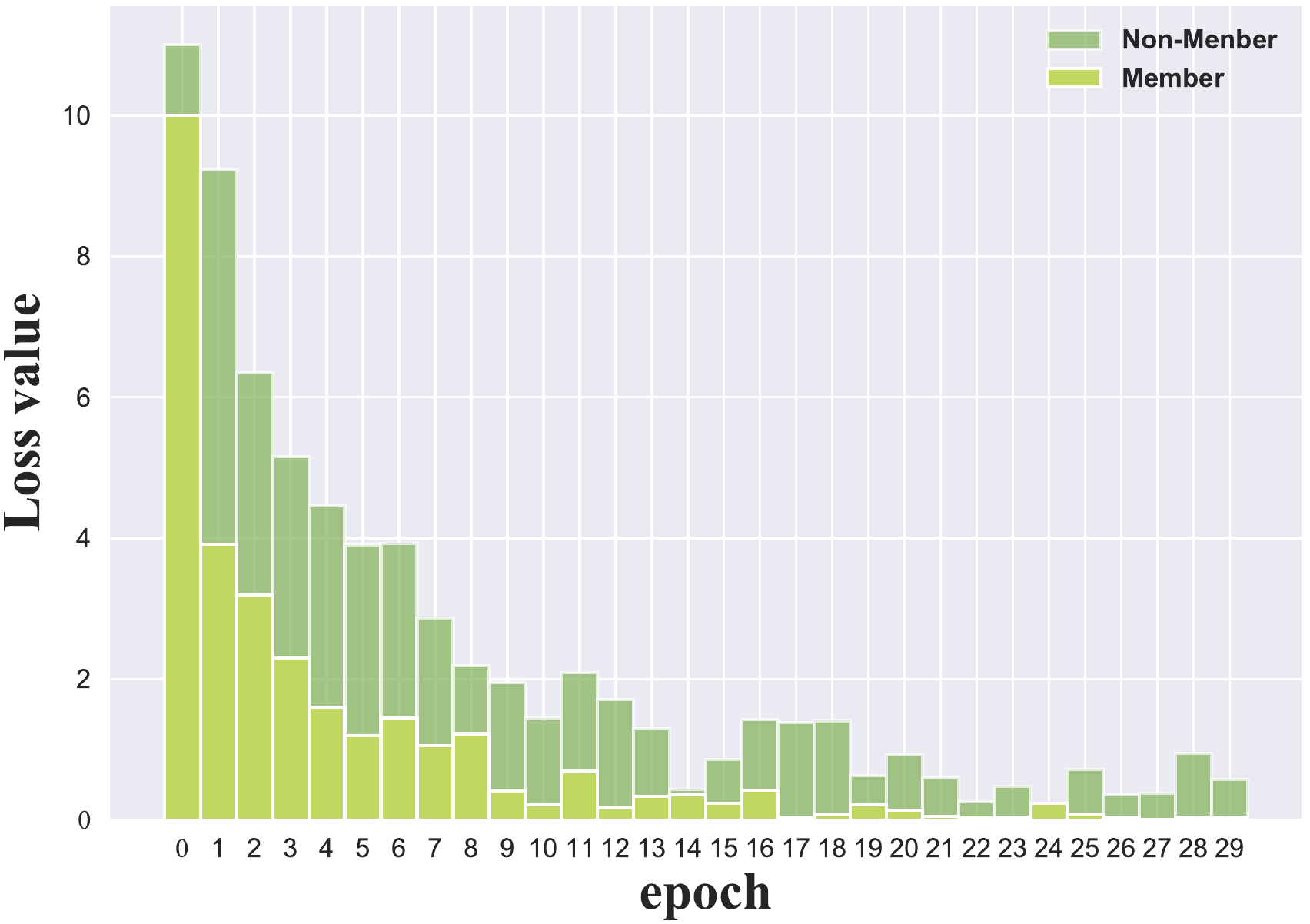}} 
\subfloat[MNIST:Shadow Model]{
\label{fig:subfig:b}
\includegraphics[scale=0.243]{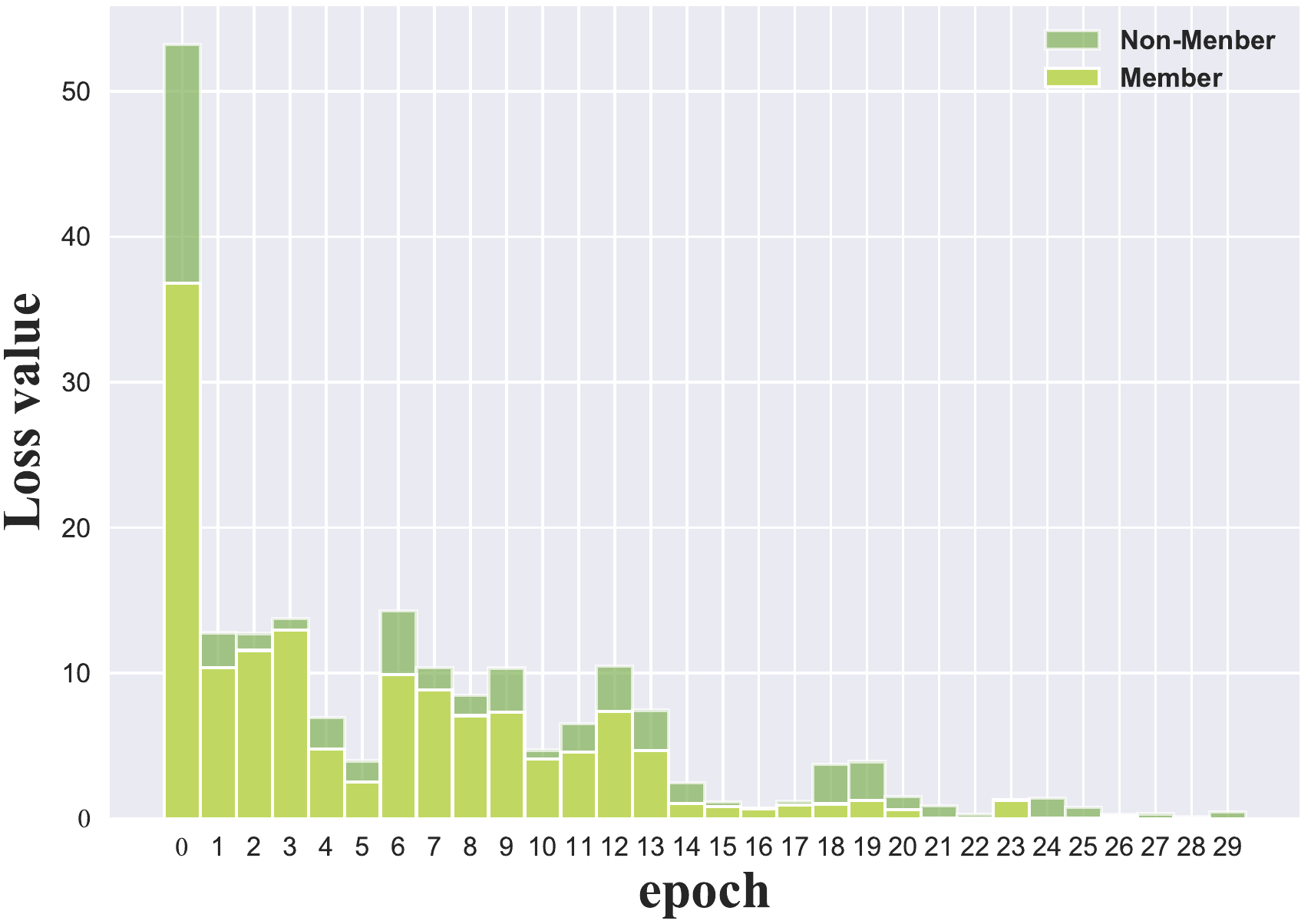}}
\subfloat[CIF10:Modify Shadow Model]{
\label{fig:subfig:c}
\includegraphics[scale=0.243]{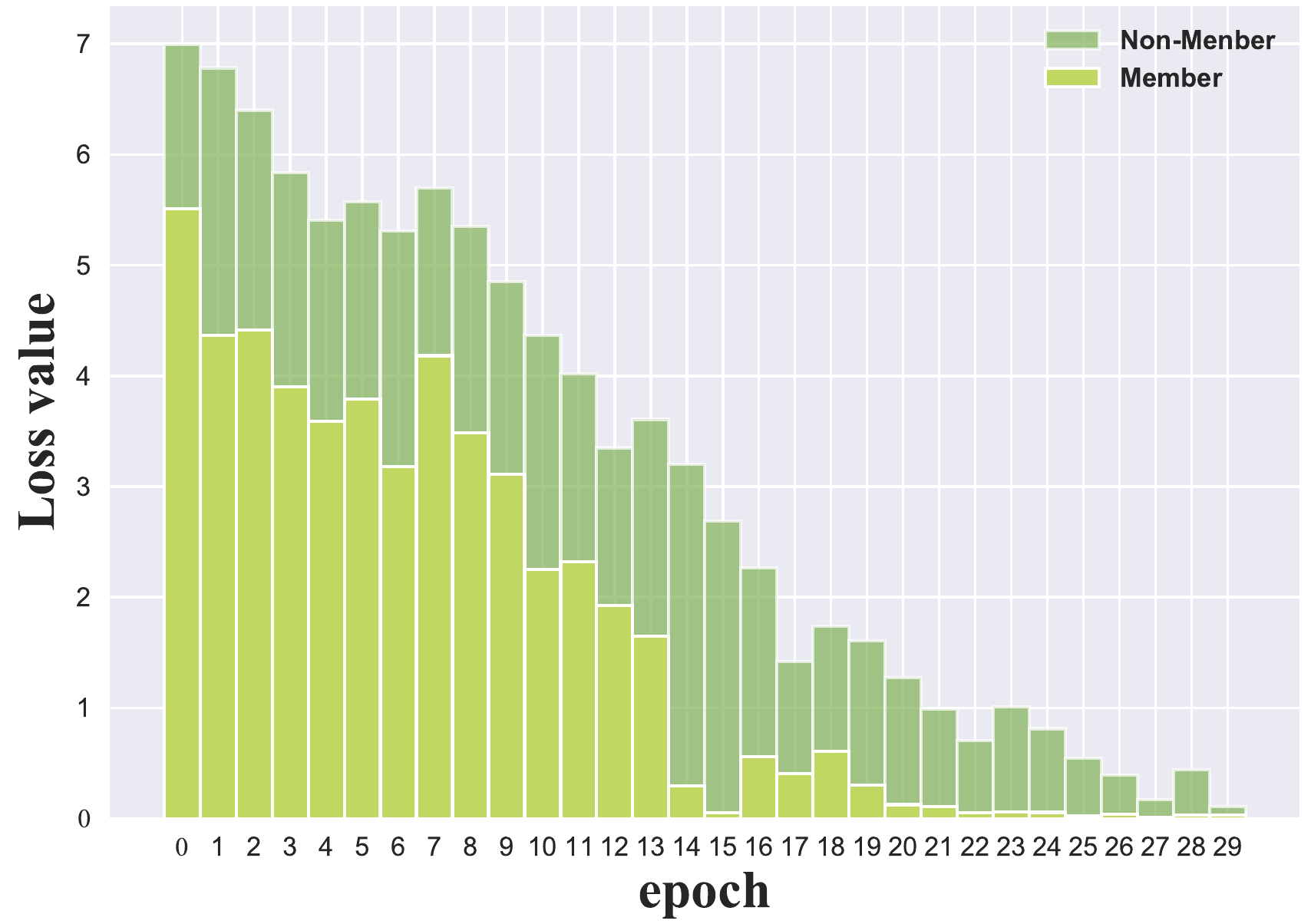}}
\subfloat[CIF10:Shadow Model]{
\label{fig:subfig:d}
\includegraphics[scale=0.243]{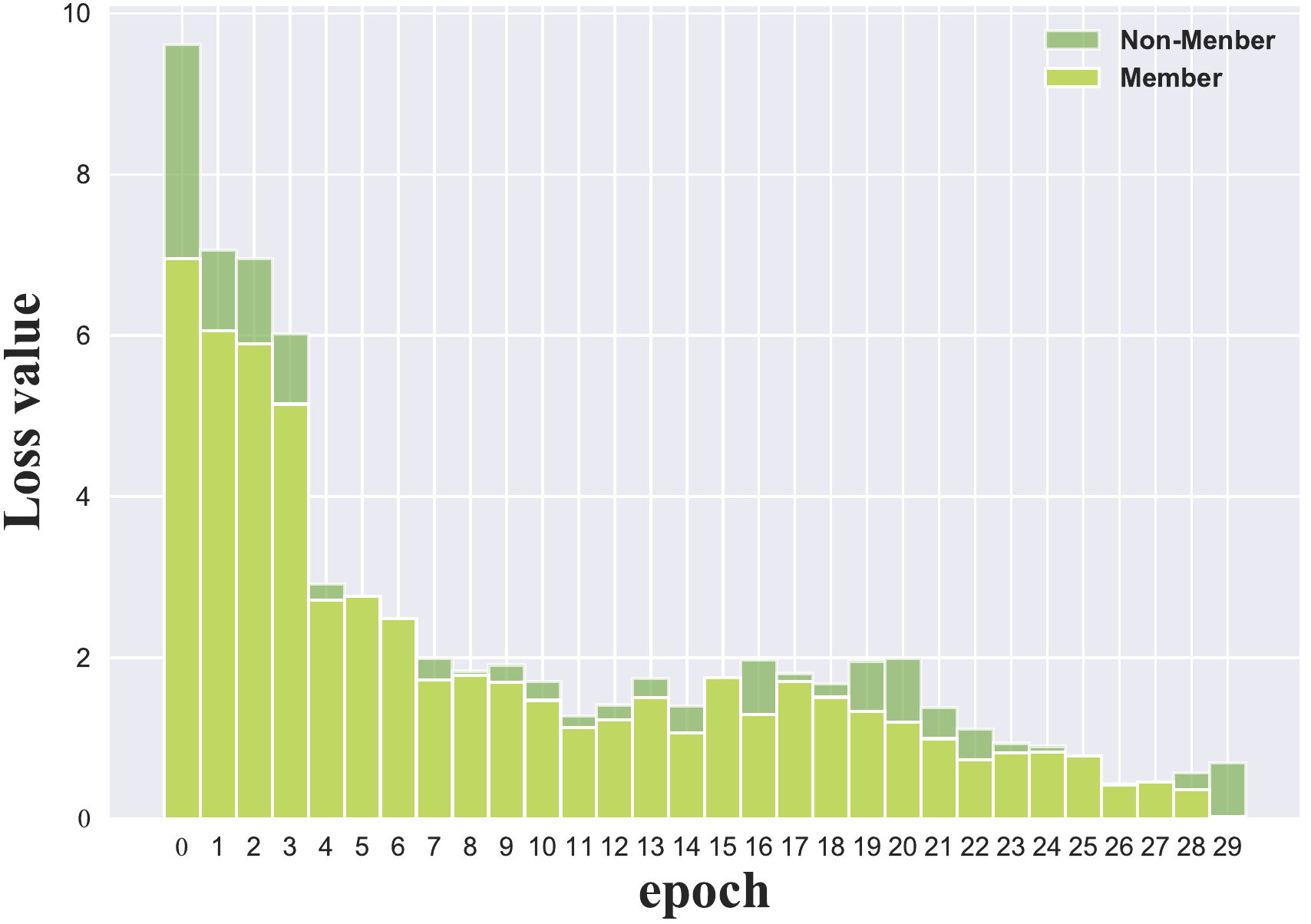}}
\caption{Measure the similarity between the shadow model and the target model. We use MNIST and CIFAR-10 to measure. We are concerned about the loss generated by the shadow model on the training set of the target model. \textbf{If the shadow model is sufficiently similar the target model, the corresponding loss will distinguish between members and non-members}. The experimental results show that the shadow model under our setting has a remarkable degree of discrimination between the member and non-member data of mnist and cifar10, indicating the effectiveness and feasibility of the shadow model imitating the target model.}
\label{fig:losscpmpare}
\end{figure*}

Fig.\ref{fig:losscpmpare} further shows the loss distributions of the target model members and non-member samples(MNIST and CIFAR10) calculated on the shadow models (the reconstructed shadow model and the unreconstructed shadow model). Although neither member nor non-member samples are used to training the shadow model, we still observe a specific difference between their losses.As shown in Fig.\ref{fig:subfig:a} and Fig.\ref{fig:subfig:c}, the difference in the l-Leaks shadow model is more prominent.This shows that our intuition is correct, and the shadow model has learned the behavior of the target model.

 To further prove that the shadow model has learned the knowledge and behavior of the target model, taking MNIST as an example. We try to delete all instances of the number $2$ from $D_{shadow}^{train}$,  so from the perspective of the shadow model, $2$ is a mythical number that has never appeared before. As shown in Tab.\ref{nolabel2}, the shadow model only produced $176$ test errors in our testing process, of which \textbf{112} were on the $1032$ $2's$ in the test set. Most of the errors are caused by a low learning bias. If this deviation increases by $4$, the shadow model will generate $89$ errors, \textbf{75} of which are on $2's$. Our shadow model achieved a \textbf{88.7\%} correct rate with the right bias in Test $2$, although it has never been seen before in training. 
 
\begin{table}[]
\scalebox{0.91}{
\begin{tabular}{ccccc}
\toprule[1.5pt]
      Shadow Model           & No 2 &  Wrong & Right Bias & Wrong \\ \hline
l-Leaks      & 82.5\%          & \textbf{112}/1032                                                  & 88.7\%             & \textbf{75}/1032           \\ 
 Prior & 10.4\%                &929/1032                                                           &-                    &-                   \\ 
\bottomrule[1.2pt]
\end{tabular}}
\caption{Shadow model performance on unseen numbers. Our shadow model achieves a prediction accuracy of \textbf{82.5\%} on unseen data because of the knowledge of the target model on the number 2. The original shadow model is insensitive to unseen numbers with only \textbf{10.4\%} accuracy.}
\label{nolabel2}
\end{table}
\subsection{Evaluation on MI-Softmax}
 Our method involves processing logits, and we compared the attack effects before and after processing. Our comparative experiment focuses on three scenarios: original shadow model; shadow model build with logits processed by softmax; shadow model build logits processed by MI-Softmax. We tested our attack effect on $5$ datasets. 
 
 As shown in Tab.\ref{MIinfluence}. Experimental results show that the impact of the original shadow model is the worst. The effect of softmax processing is better than the former. The shadow model processed by MI-Softmax has the best attack effect, especially in the image classification task. We can see that the attack success rate on the MNIST dataset has increased by \textbf{29\%}. In the Cifar10 dataset, it has risen by about \textbf{14\%}.
\begin{table}[]
\scalebox{0.92}{
\begin{tabular}{ccccc}
\toprule[1.5pt]
Dataset & General & Softmax & MI-Softmax & Compare \\ \cline{2-5} 
                         & $F_1S$     & $F_1S$     & $F_1S$        & $F_1S$        \\ \hline
Purchase                 & 0.77    & 0.79    & 0.85       & +8\%       \\ 
Texas                    & 0.74    & 0.75    & 0.80       & +6\%      \\ 
MNIST                    & 0.62    & 0.65    & 0.91       & \textbf{+29\%}      \\ 
CIFAR10                  & 0.80    & 0.81    & 0.94       & \textbf{+14\%}     \\
CIFAR100                 & 0.83    & 0.85    & 0.95       & +12\%  \\
\bottomrule[1.2pt]
\end{tabular}}
\caption{MIA \textbf{F1-Score} under different construction conditions of the shadow model. Test five different datasets under three cases. The comparison experiment shows that the shadow model built by MI-Softmax has the highest attack success rate, especially in image dataset MNIST, an increase of \textbf{29\%}.}
\label{MIinfluence}
\end{table}
\subsection{Evaluation on Different Network Structures}
Another advantage of l-Leaks is to build shadow models through learning knowledge and the structure of the shadow model is not restricted. So we can use different network structures to act as the shadow model. 

This experiment section set the shadow model to three different structures: the classic CNN network-VGG, our own locally implemented NN (CNN+FC), FC (Fully Connect Network). We didn't follow the previous work\cite{li2020membership} that continuously improved the parameter amount of the shadow model. Instead, we reduce the shadow model parameters, and the final attack effect is still strong. As shown in Tab.\ref{networks}, we tested on 5 datasets. As the model parameters decrease, the accuracy of the attack falls within a \textbf{tolerable} range. Tolerable means that we save calculation cost and time under the premise of reducing the number of parameters. Our price is that the attack accuracy dropped by an average of $2.2\%$ on the $5$ datasets.The reduced attack accuracy rate is still higher than the previous work\cite{salem2019ml}\cite{li2020membership}. Therefore, our accuracy rate has received a minor impact. At the same time, the calculation cost is reduced. Compared with the previous work, our work is economical.

Summarizing the above experiments, our shadow model imitates the behavior of the target model by learning the logits of the target model. Experiments prove that the performance of the shadow model we constructed through learning is excellent, and the effect is robustness.

\begin{table}[]
\scalebox{0.92}{
\begin{tabular}{cccccc}
\toprule[1.5pt]
NetWork & Purchase & Texas & MNIST & CIF10 & CIF100 \\ \cline{2-6}
                         & $F_1S$      & $F_1S$   & $F_1S$   & $F_1S$   & $F_1S$    \\ \hline
VGG                      & 0.86    & 0.79  & 0.91  & 0.95  & 0.95   \\ 
NN                       & 0.85     & 0.80  & 0.91  & 0.94  & 0.95   \\ 
FC                       & 0.83     & 0.79  & 0.88  & 0.94  & 0.91   \\ \hline
Compare                       & -3\%     & \textbf{0\%}  & -3\%  & \textbf{-1\%}  & -4\%   \\ 
\bottomrule[1.2pt]
\end{tabular}}
\caption{MIA \textbf{F1-score} of shadow model under different network structures. In the complex network VGG and the simple network FC, our attack performance is reduced around $\textbf{2\%}$, which is acceptable. It is precise because our attack relies on the shadow model's ability to learn from the target model. The shadow model we get is similar to the target model, so the shadow model has greater confidence in the member data of the target model.}
\label{networks}
\end{table}
\section{Conclusion and Future Work}
\label{sec:experiment}
\subsection{Conclusion}
In this paper, we systematically investigate the L-leaks membership inference attack with logits to solve the problem of shadow model construction. We propose an attack for constructing shadow models based on theory, namely \textbf{l-Leaks}. For l-Leaks, she uses the target model as an oracle to inquire her shadow dataset and get the output of the target model-logits. We proposed \textbf{MI-Softmax} to process logits and modify the loss function of the shadow model to reduce the logits distribution of the shadow model and target model, the training of the shadow model based on it.

Our work demonstrates that our attack achieves strong performance, and a theoretical basis can support the construction of the shadow model. In the future, we can use the different knowledge of the target model to construct the shadow model through various learning methods. We believe our work can serve as a stepping stone for more advanced attacks in the future. 
\subsection{Future Work}
The overfitting of the target ML model is the main factor for the success of MIAs.The overfitting phenomenon of the ML model is usually caused by two reasons, the high model complexity and the limited size of the training dataset. Deep learning models learn effectively from big data and use many epochs to repeat training on the same instance, so the model easily memorizes the training instance. To the above problems, our next step is to design a privacy detection framework for deep learning models, use MIA to evaluate the degree of model leakage in actual production, and process the model's training data to reduce the degree of model overfitting. By MIA detecting and processing over-fitting, the model's performance and the degree of leakage of the model can reach a dynamic equilibrium state during the training process. Ultimately, prevent data leakage attacks fundamentally.
{\small
\bibliographystyle{ieee_fullname}
\bibliography{egbib}
}

\end{document}